\documentclass[conference]{IEEEtran}
\usepackage{graphics} 
\usepackage{epsfig} 
\usepackage{amsmath} 
\usepackage{amssymb}  
\usepackage{float}

\begin{document}

\title{A Novel Propulsion Method of Flexible Underwater Robots}

\author{\IEEEauthorblockN{Jun Shintake, Aiguo Ming and Makoto Shimojo}
\IEEEauthorblockA{Department of Mechanical Engineering and Intelligent Systems\\
The University of Electro-Communications\\
1-5-1 Chofugaoka, Chofu-shi, Tokyo 182-8585, Japan\\
Email: shintake@rm.mce.uec.ac.jp, ming@mce.uec.ac.jp, shimojo@mce.uec.ac.jp}
}

\maketitle

\begin{abstract}
This paper presents aims at mobility improvement of flexible underwater robots.
For this purpose, a novel propulsion method using planar structural vibration pattern is proposed, and tested on two kinds of prototypes.
The result of experiments showed the possibility of the movements for multiple directions:
forward, backward, turn, rotation, drift, and their combination.
These movements are achieved by only one structure with two actuators.
The results also indicated the possibility of driving using eigenmodes since movements were concentrated on low driving frequency area.
To investigate the relation between movement and structural vibration pattern, we established a simulation model.
\end{abstract}

\section{Introduction}
In recent years, underwater robots come to be used for
exploration, observation, research, and salvage.

Underwater creatures are capable of high movement performance
in water and they have various forms, so it can be said
that underwater robot design based on the movement mechanism
of the underwater creatures is an effective method.

Over the past few years, many researches have been
done on developing fish type bio-mimetic robot as one form of underwater
robot based on creatures' swimming mechanism [1]-[3].
Most of the fish type robots developed up to now have complicated mechanisms using motors combined with cranks
etc., and the control of the robots is also complicated [4], [5].
The movement of the robots is like a mechanical one, it is considered that the performance of the robots is limited due to lack of flexibility and fluency[6].
Therefore, taking a creature-like flexibility to the whole structure of the robots have attracted attention recently.
For this, applying the soft actuators as an artificial muscle to the robots have been attempted[7]-[10].

We also have studied such flexible underwater robots using piezoelectric fiber composites as a soft actuator [11], [12].
The new type of piezoelectric fiber composite we noticed has flexibility, large displacement and high mechanical force.
Compared with conventional artificial muscles, piezoelectric fiber composite can be used for composing powerful underwater robots with a very simple and compact structure.
For example, an underwater robot can be made to behave like a genuine fish with only one piece of piezoelectric fiber composite and a thin structural plate.
For such a structure, a flexible and fluent motion can be achieved and it has reduced fluid resistance and adaptability for narrow space.
The piezoelectric fiber composite can be used for actuating, sensing and energy harvesting.
Therefore, there is possibility to realize an intelligent underwater robot with intelligent motion control in the future.

We developed fish like flexible underwater robots by meandering propulsion,
which is capable of fast motion(0.32m/s, 2.9BL/s) and planar motion [11].
Also, we have tried to improve the mobility performance of these robots,
the robot using caudal fin propulsion was developed and more fast motion (0.72m/s, 4.3BL/s) and pitch/roll motion were demonstrated up to now [12].

This paper aims further improvement of the mobility performance of flexible underwater robots. Different with the conventional propulsion methods of the underwater robots up to now,  we propose a novel propulsion method utilizing both static and dynamic deformation of a flexible planar structure. By actuating the planar structure in two dimensions, various propulsive motions can be generated. And the realization of various robotic movements, such as movements with multiple DOFs, holonomic and non-holonomic movements, can be expected.
First, the basic feature of piezoelectric fiber composite is introduced in section II.
And a novel propulsion method is described in section III.
Two kinds of robots based on the propulsion method are shown in section IV, and the result of experiments is shown in section V.
The discussion using FEM simulation on the results is described in section VI. Section VII summarizes the results.

\section{Piezoelectric fiber composite}
\begin{figure}[H]
  \begin{center}
    \resizebox{85mm}{!}{\includegraphics{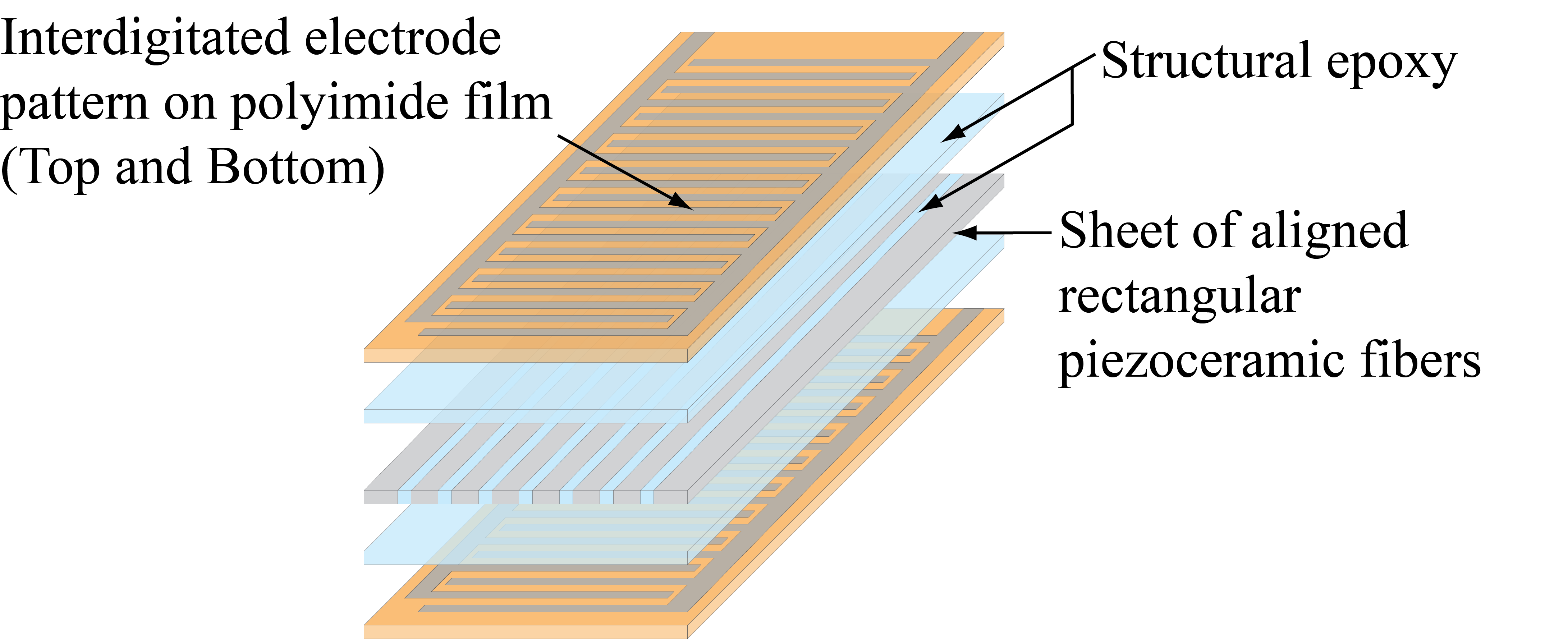}}
    \caption{Structure of Macro Fiber Composite}
    \label{fig:MFC_structure}
  \end{center}
\end{figure}

We used one of the typical piezoelectric fiber composites, Macro Fiber Composite (MFC [13]) for this study.
Fig.\ref{fig:MFC_structure} shows the structure of MFC.
With structural features such as rectangular piezoceramic fibers and interdigitated
electrode pattern on the polyimide film, this structure has
better flexibility and large strain compare to conventional piezoelectric actuators.
When a voltage is applied, this actuator expands and contracts in the direction of the fibers.
If the actuator is attached to thin structure, it will realize the bending motion like a swimming motion of creatures,
and also gives a resonant behavior with large displacement.

\section{Propulsion method}
\subsection{Concept}
\begin{figure}[H]
  \begin{center}
    \resizebox{85mm}{!}{\includegraphics{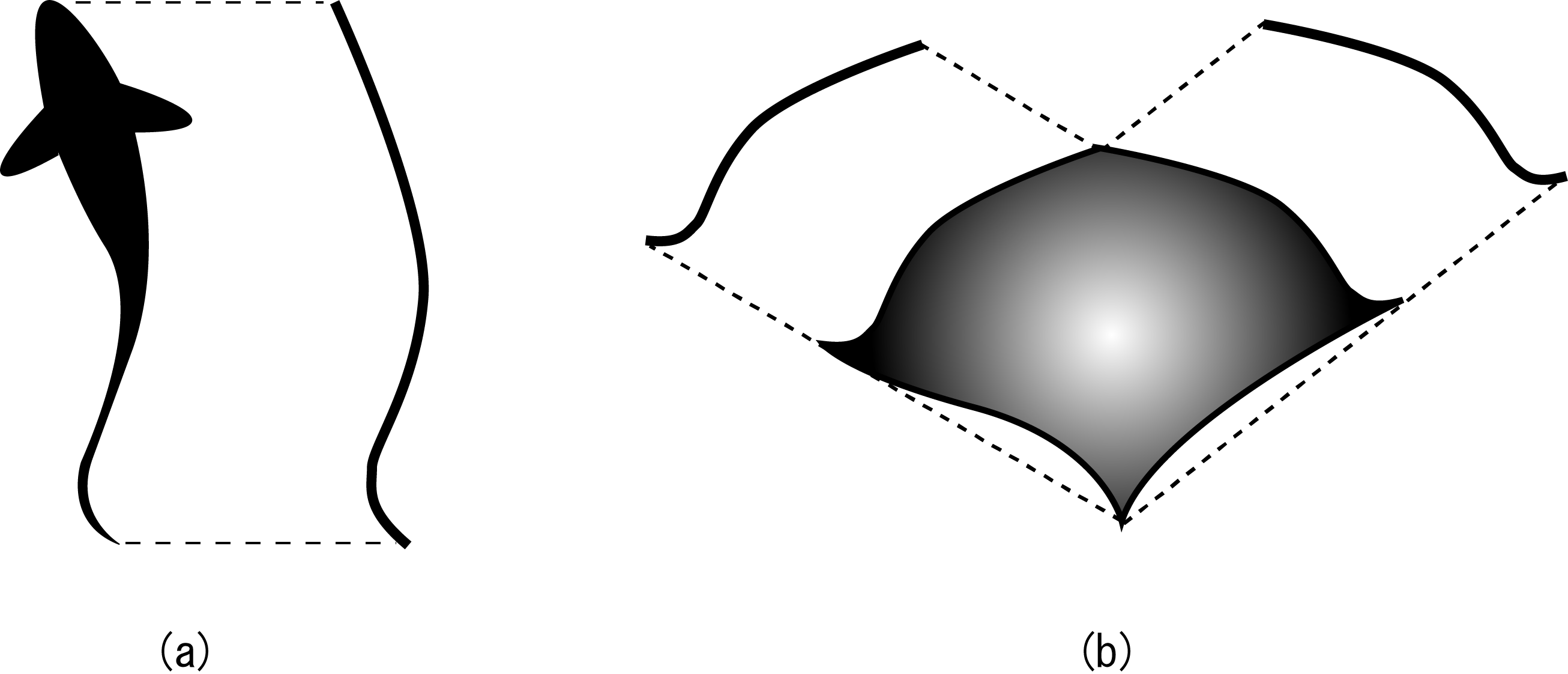}}
    \caption{Robotic motions, (a): one-dimensional mode, (b): two-dimensional mode.}
    \label{vib_mode}
  \end{center}
\end{figure}

In our previous work, robotic motion of one-dimensional vibration mode like Fig.\ref{vib_mode}(a) has been considered.
The propulsion method to be proposed is that, extending robotic motion from one-dimensional to two-dimensional like Fig.\ref{vib_mode}(b).

\vspace{-1mm}
\vspace{-2mm}

\begin{figure}[H]
  \begin{center}
    \resizebox{70mm}{!}{\includegraphics{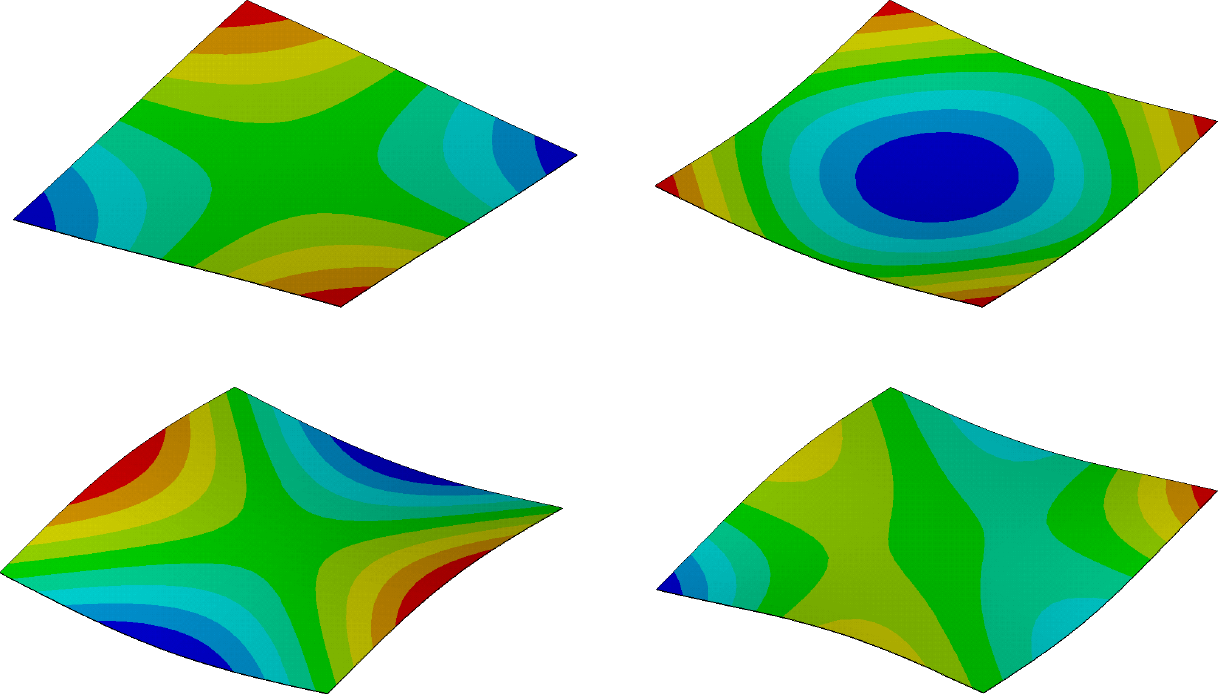}}
    \caption{Eigenmodes of rectangular plate}
    \label{modes}
  \end{center}
\end{figure}

\begin{figure}[H]
  \begin{center}
    \resizebox{85mm}{!}{\includegraphics{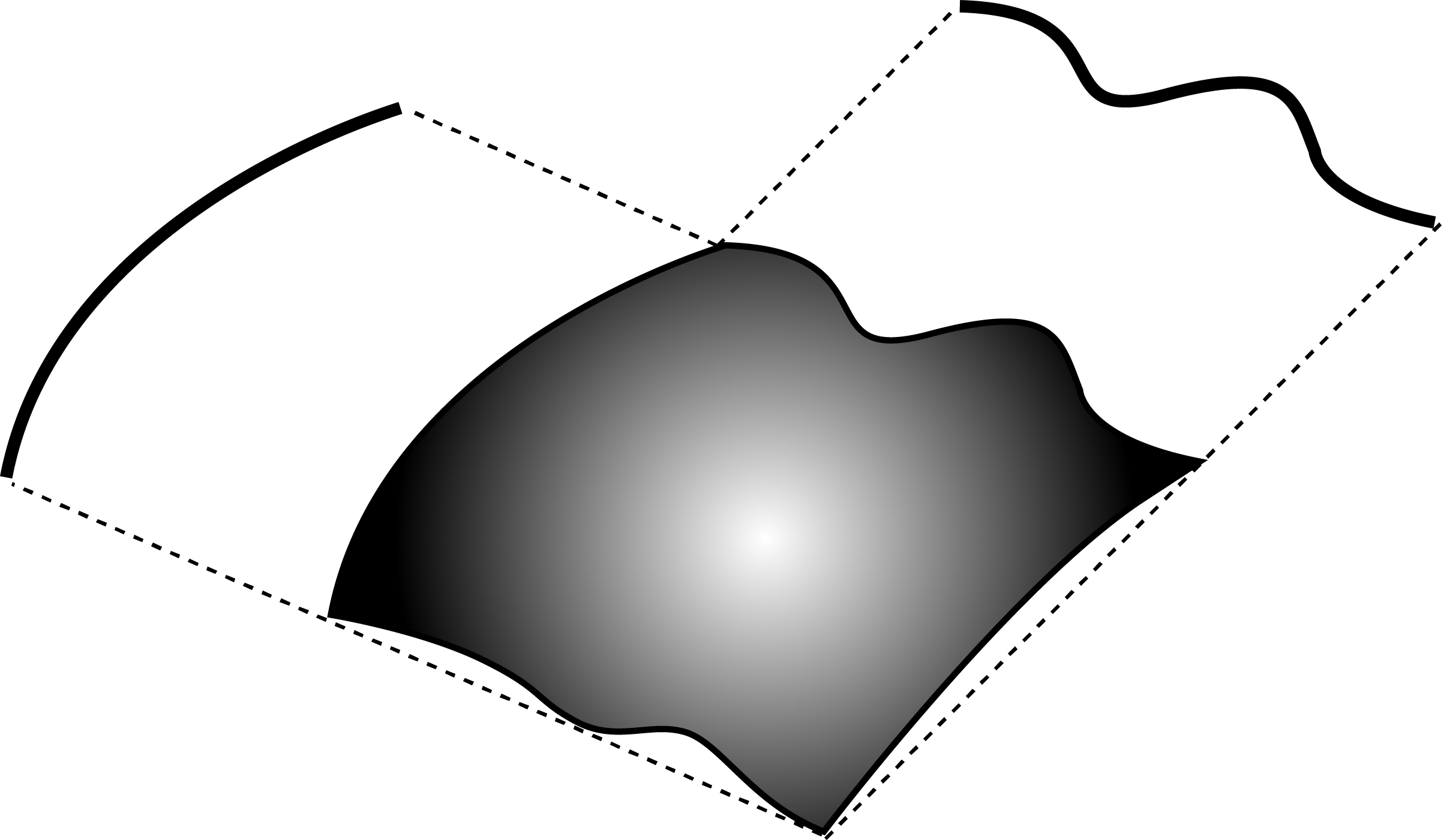}}
    \caption{The cross-mode}
    \label{cross}
  \end{center}
\end{figure}

This method can be used in planar structure such as thin plate and membrane, because there exist various vibration modes in this kind of structure.
Fig.\ref{modes} shows typical eigenmodes of a rectangular plate.
If these modes are excited in the water, there is possibility to generate fluid-flows for particular direction movement corresponding to each mode.
Moreover, in planar structure, it is considered that we can get more vibration patterns by combining several modes generated by multiple actuations on the structure, shown as Fig.\ref{cross}.
In this study, we call this crossing actuation as "cross-mode".

\vspace{-1mm}
\vspace{-2mm}

\vspace{-2mm}
\vspace{-2mm}

\subsection{Actuation method}
It is considered that the cross-mode should be complicated,
and it is difficult to control by actuation for general shape structures.
One of the methods to solve this problem is using the degenerate mode to control the cross-mode.
It is known that the degenerate mode appears on symmetric shapes such as rectangular and circular,
where there are two modes with one eigenfrequency.

\vspace{-2mm}
\begin{figure}[H]
  \begin{center}
    \resizebox{50mm}{!}{\includegraphics{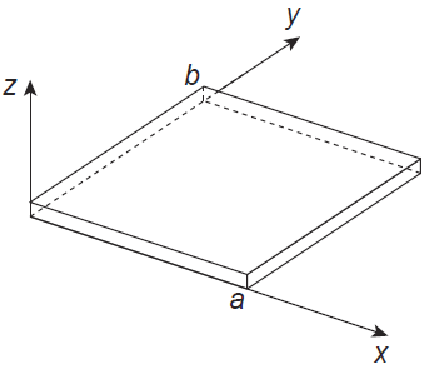}}
    \caption{Coordinate system of a rectangular plate}
    \label{coo}
  \end{center}
\end{figure}
\vspace{-2mm}

In Fig.\ref{coo}, the equation of the vibration of a thin plate as a planar structure is

\begin{equation}
D\left(\frac{{\partial}^{4}w}{\partial{x^4}}+2\frac{{\partial}^{4}w}{\partial{x^2}\partial{y^2}}+\frac{{\partial}^{4}w}{\partial{y^4}}\right)
+\mu\frac{{\partial}^{2}w}{\partial{t^2}}-q_{z}(x,y,t)=0
\label{eq51}
\end{equation}

\begin{equation}
D=\frac{Eh^{3}}{12(1-\nu^{2})}
\label{eq52}
\end{equation}

Where $w$ is displacement of direction $z$ at planar position of $(x,y)$, $D$ is bending stiffness, $E$ is young's modulus, $h$ is thickness of plate, $\mu$ is mass per unit volume, and $\nu$ is Poisson's ratio.
In Eq.(\ref{eq51}), solution of free vibration ($q_z = 0$) is

\begin{equation}
w(x,y,t)=W(x,y)\cos\omega{t}
\label{eq53}
\end{equation}

By substituting Eq.(\ref{eq53}) into Eq.(\ref{eq51}), we can get equation about $W(x,y)$

\begin{equation}
D\Delta{\Delta}W-\mu\omega^{2}W=0~~~~~\Delta=\frac{{\partial}^{2}}{\partial{x^2}}+\frac{{\partial}^{2}}{\partial{y^2}}
\label{eq54}
\end{equation}

For simple support boundary condition, we can get simple solution

\begin{equation}
W(x,y)=\sin\frac{m\pi{x}}{a}\sin\frac{n\pi{y}}{b}
\label{eq55}
\end{equation}

Where $m$ and $n$ are integer, and $a$ and $b$ are length of sides. By substituting Eq.(\ref{eq55}) into Eq.(\ref{eq54}), we can get eigenfrequency

\begin{equation}
\omega_{mn}=\pi^{2}\left(\frac{m^{2}}{a^{2}}+\frac{n^{2}}{b^{2}}\right)\sqrt{\frac{D}{\mu}}
\label{eq56}
\end{equation}

Eq.(\ref{eq55}) shows the shape of eigen modes, and it is changed by value of $m$ and $n$. If the structure is square $(a = b)$, eigen modes of $(m,n)$ and $(n,m)$ are shown by

\begin{equation}
W_{mn}=\sin\frac{m\pi{x}}{a}\sin\frac{n\pi{y}}{b}
\label{eq57}
\end{equation}

\begin{equation}
W_{nm}=\sin\frac{n\pi{x}}{a}\sin\frac{m\pi{y}}{b}
\label{eq58}
\end{equation}

These two modes have same eigenfrequency, it can be calculated by

\begin{equation}
\omega_{mn}=\omega_{nm}=\pi^{2}\left({m^{2}}+{n^{2}}\right){\frac{1}{a^{2}}}\sqrt{\frac{D}{\mu}}
\label{eq59}
\end{equation}

This is called degeneracy. And Eq.(\ref{eq57}) and Eq.(\ref{eq58}) are called degenerate mode.
There is possibility of these two modes occur at once, but only one mode appears as unspecified mode shape.
It can be described by superposition of those two equations as

\begin{equation}
W=W_{mn}\cos{\gamma} + W_{nm}\sin{\gamma}
\label{eq60}
\end{equation}

\begin{figure}[H]
  \begin{center}
    \resizebox{60mm}{!}{\includegraphics{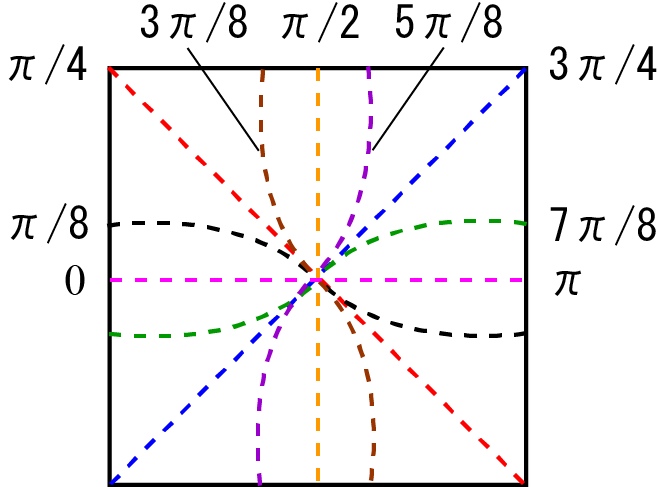}}
    \caption{Degenerate modes}
    \label{aaa}
  \end{center}
\end{figure}

Fig.\ref{aaa} shows the example of degenerate mode shapes which is obtained by changing value of $\gamma$.
In this figure, each line means a nodal line of mode shapes.
In fact, the modes change their direction or shape, determined by initial condition and vibration position.
The degenerate modes exist for any boundary conditions.
If the structure produces this kind of modes in water, it is considered that there is possibility to generate various movements corresponding to the mode shapes.
We assume that we can produce the degenerate modes by using superposition of different actuations. That is, the cross-mode.

\begin{figure}[H]
  \begin{center}
    \resizebox{85mm}{!}{\includegraphics{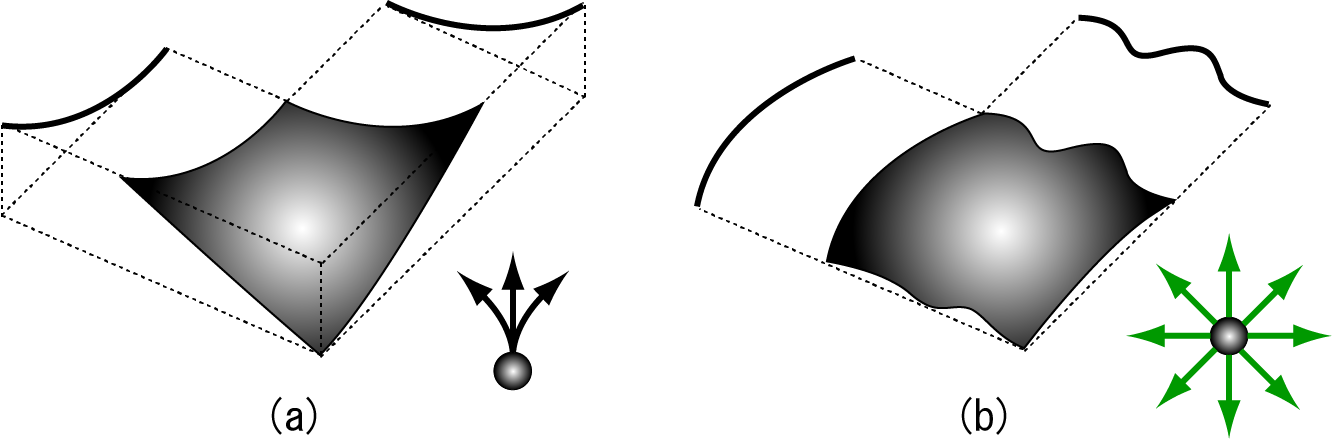}}
    \caption{Actuation modes and movement types}
    \label{types}
  \end{center}
\end{figure}

When we put actuators to appropriate position such as leading edge, flapping motion as a creature-like, non-holonomic movement can be realized by synchronous actuation as shown in Fig.\ref{types}(a).
Also, asynchronous actuation, movements for multiple directions, holonomic movement can be realized as shown in Fig.\ref{types}(b).
That is, the propulsion method we proposed has possibility to realize both holonomic and non-holonomic movements by different actuation, using only one structure.

The features of the method are as follows.

\vspace{1mm}
\begin{itemize}
\item Many directions of movements and propulsive DOFs are available by few actuators.
\item Only flexible structures can be used for this method, and soft actuators are suitable for direct actuation on the structure.
\item The structure is simple and it is easy to downsize.
\end{itemize}
\vspace{1mm}

To investigate the feasibility of our proposal, we designed the prototypes of underwater robot by considering following conditions.

\vspace{1mm}
\begin{itemize}
\item Symmetric shape for degenerate mode.
\item Right-angled actuator location for the cross-mode and flapping motion.
\end{itemize}
\vspace{1mm}

\section{Prototypes of underwater robot}
\begin{figure}[H]
  \begin{center}
    \resizebox{65mm}{!}{\includegraphics{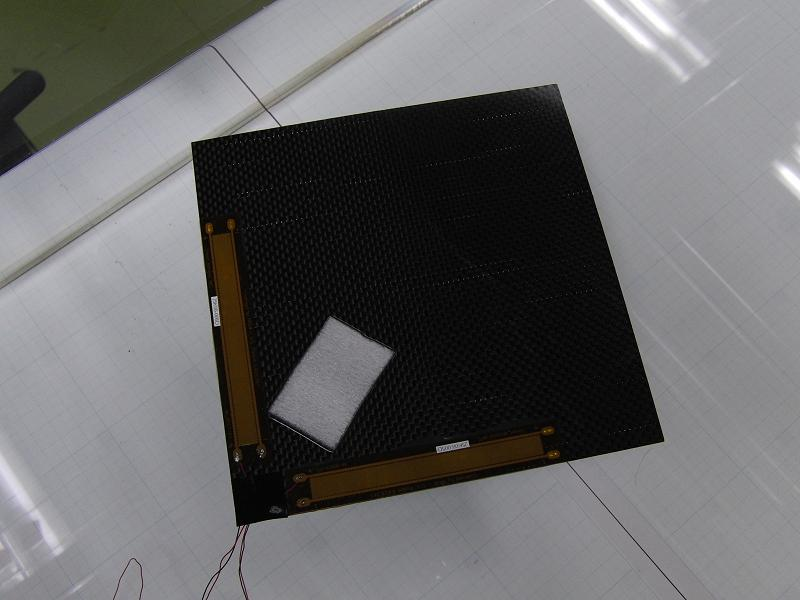}}
    \caption{Rectangular robot}
    \label{rec}
  \end{center}
\end{figure}

\begin{figure}[H]
  \begin{center}
    \resizebox{65mm}{!}{\includegraphics{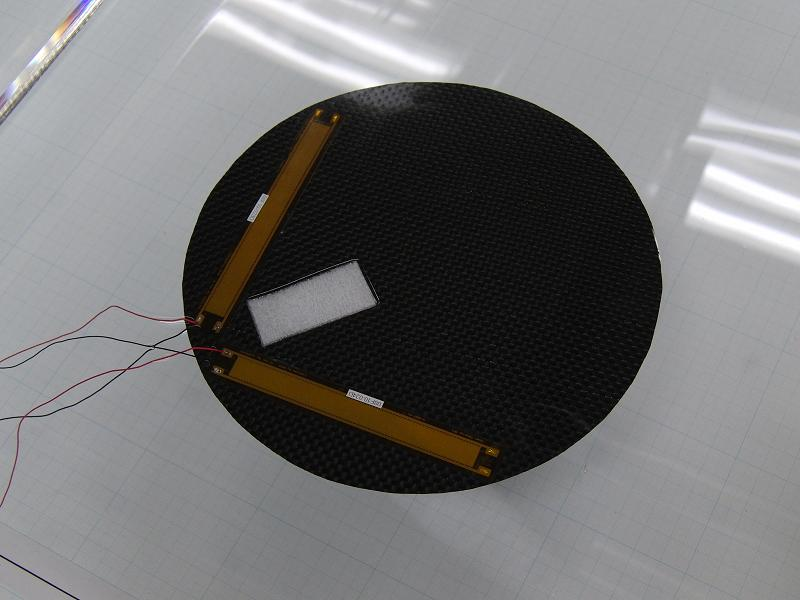}}
    \caption{Circular robot}
    \label{circle}
  \end{center}
\end{figure}

We developed two kinds of robots, both of them have symmetric shape for degenerate mode,
and also have right-angled actuator location to achieve the cross-mode.
One is rectangular shape shown in Fig.\ref{rec}, and Tab.1 shows the specification.
The other is circular shape shown in Fig.\ref{circle}, and Tab.2 shows the specification.
These robots have a float at the top of their body to balance.

\begin{table}[H]
\begin{center}
\vspace{-1mm}
\caption{Specification of rectangular robot}
\begin{tabular}{cc}
\hline
Length & 226mm \\ 
Width & 226mm \\ 
Length of side & 160mm \\ 
Weight & 13.5g \\ 
Actuator overall dimensions & 101mm$\times$13mm\\
Actuator active area & 85mm$\times$7mm\\
Material for adhering & Carbon plate 0.2mm\\
Adhesion bond & Epoxy Adhesives DP460\\
\hline
\end{tabular}
\label{tb:spec}
\end{center}
\end{table}


\begin{table}[H]
\begin{center}
\vspace{-1mm}
\caption{Specification of circular robot}
\begin{tabular}{cc}
\hline
Radius & 80mm \\
Weight & 9.68g \\ 
Actuator overall dimensions & 101mm$\times$13mm\\
Actuator active area & 85mm$\times$7mm\\
Material for adhering & Carbon plate 0.2mm\\
Adhesion bond & Epoxy Adhesives DP460\\
\hline
\end{tabular}
\label{enspe}
\end{center}
\end{table}

\section{Experimental result}
The experiments were conducted in a water tank which is filled by insulated fluid (Fluorinert FC-3283) for safety.
At first, we confirmed non-holonomic movements such as forward movement and forward turning movement.
The forward movement appeared at synchronous drive of the two actuators, and the forward turning movement appeared at driving one of the two actuators.
Next, we tested and observed robotic movements to find holonomic movements by changing the driving condition such as frequency, phase difference between the two actuators, and voltage range.
We used sine wave as an input wave form, and typical experiment condition is as follows.

\vspace{-2mm}

\begin{table}[H]
\begin{center}
\begin{tabular}{ll}
{\bf (Experiment condition)}& \\
Voltage range&$-500$V$\sim+1500$V\\
Frequency&1Hz$\sim$\\
Phase difference & 0deg$\sim$\\
Input wave form& Sine wave\\
\end{tabular}
\end{center}
\end{table}

\vspace{-2mm}

\subsection{Rectangular shape robot}
\subsubsection{Backward movement}
\begin{figure}[H]
  \begin{center}
    \resizebox{85mm}{!}{\includegraphics{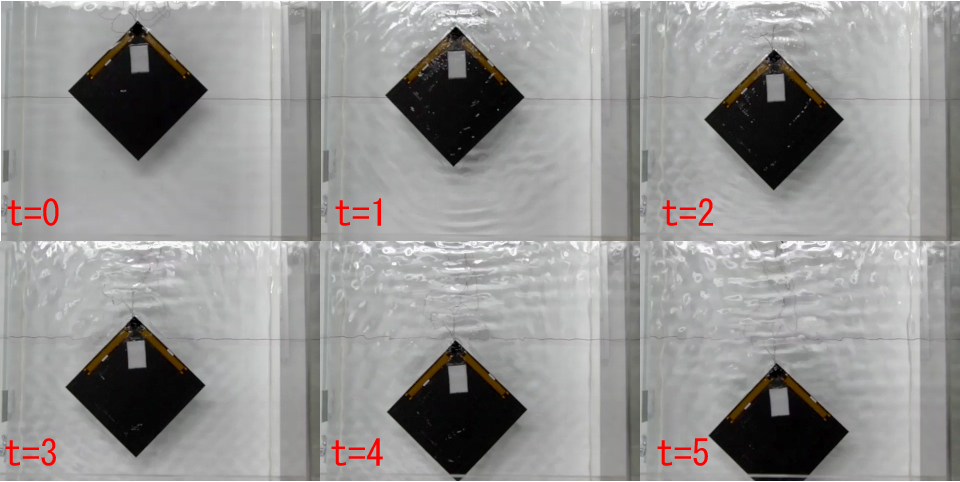}}
    \caption{Backward movement}
    \label{back}
  \end{center}
\end{figure}

Backward movement was observed at frequency of 6Hz with 0deg phase difference.
Fig.\ref{back} shows the backward movement by the time $t$ (sec).
And, same movement was observed when the voltage range was changed to $+1000$V$\sim+1500$V at driving frequency of 6Hz$\sim$10Hz.

\begin{figure}[H]
  \begin{center}
    \resizebox{85mm}{!}{\includegraphics{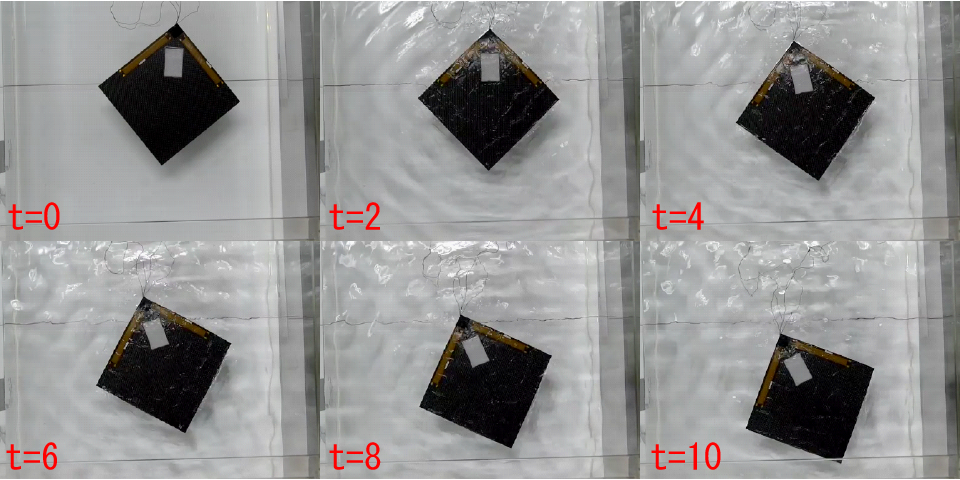}}
    \caption{Backward turning movement}
    \label{ks}
  \end{center}
\end{figure}

At this frequency of 6Hz, direction of robotic movements were changed by phase differences.
Fig.\ref{ks} shows the backward turning movement observed at 15deg phase difference.
When the phase difference was reversed to $-$15deg, the direction of the movement was also reversed.
For other phase differences, turning rate of robotic movements were changed and translational.
When the phase difference was 180deg, forward movement appeared.

\subsubsection{Rotational movement}
\begin{figure}[H]
  \begin{center}
    \resizebox{85mm}{!}{\includegraphics{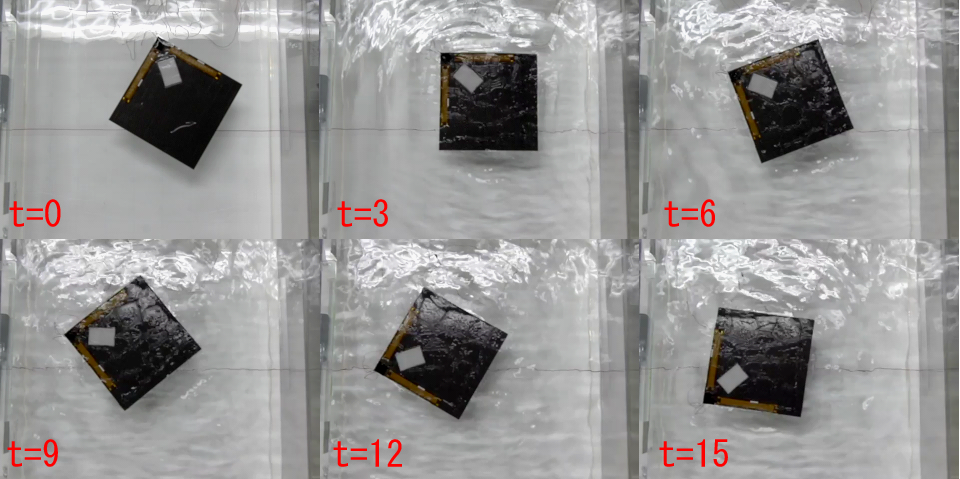}}
    \caption{Rotational movement}
    \label{kaiten}
  \end{center}
\end{figure}

Rotational movement was observed at frequency of 5Hz with 90deg phase difference, shown in Fig.\ref{kaiten}.
When the phase difference was reversed, the movement was also reversed.
The robot was static at 0deg and 180deg phase differences.
For other phase differences, forward turning movements appeared.

\subsubsection{Translational movements}
\begin{figure}[H]
  \begin{center}
    \resizebox{85mm}{!}{\includegraphics{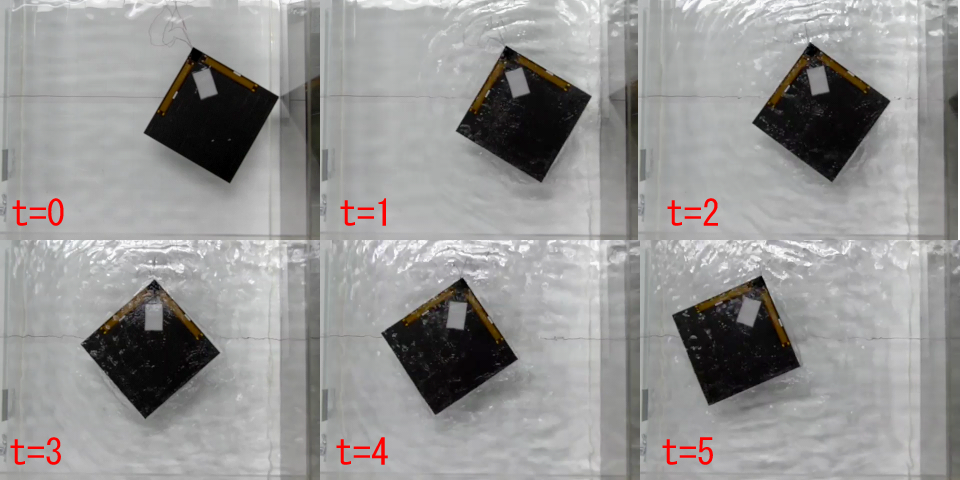}}
    \caption{Translational movement for sideways}
    \label{yoko}
  \end{center}
\end{figure}

Translational movement for sideways was observed at frequency of 8Hz with 60deg phase difference, shown in Fig.\ref{yoko}.
When the phase difference was reversed, the movement was also reversed.

\begin{figure}[H]
  \begin{center}
    \resizebox{85mm}{!}{\includegraphics{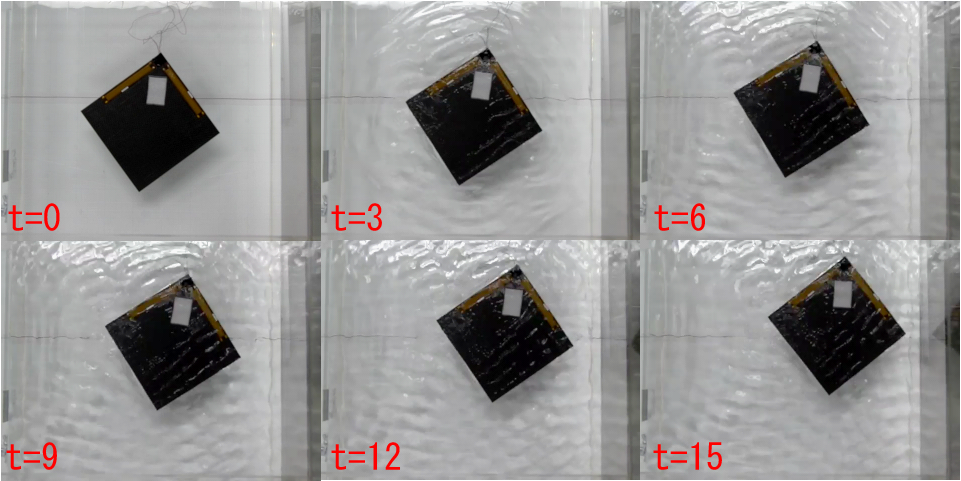}}
    \caption{Translational movement for diagonal direction}
    \label{naname}
  \end{center}
\end{figure}

At the frequency, the robot was static at 0deg phase difference.
For other phase differences, turning movements and translational movements were observed.
Fig.\ref{naname} shows the translational movement for diagonal direction, which was observed at 120deg phase difference.
When the phase difference was 180deg, forward movement appeared.

\subsection{Circular shape robot}
\subsubsection{Backward movement}
\begin{figure}[H]
  \begin{center}
    \resizebox{85mm}{!}{\includegraphics{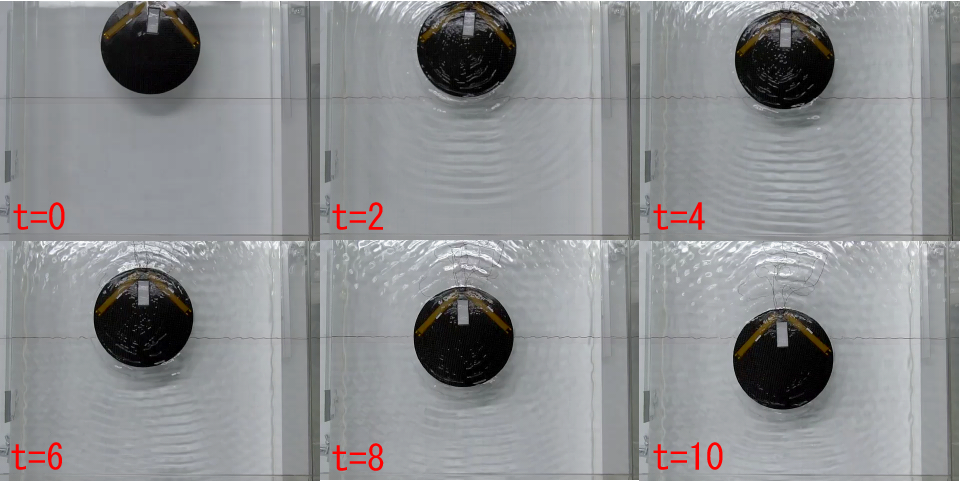}}
    \caption{Backward movement}
    \label{minikou}
  \end{center}
\end{figure}

Backward movement was observed when the voltage range was changed to $+1000$V$\sim+1500$V which same to the case of recutanglar shape, at driving frequency of 8Hz$\sim$10Hz with 0deg phase difference, as shown in Fig.\ref{minikou}.
In this frequency range, direction of robotic movements were changed by phase differences.
Backward movement also appeared when the phase difference was 180deg.

\subsubsection{Rotational movement}
\begin{figure}[H]
  \begin{center}
    \resizebox{85mm}{!}{\includegraphics{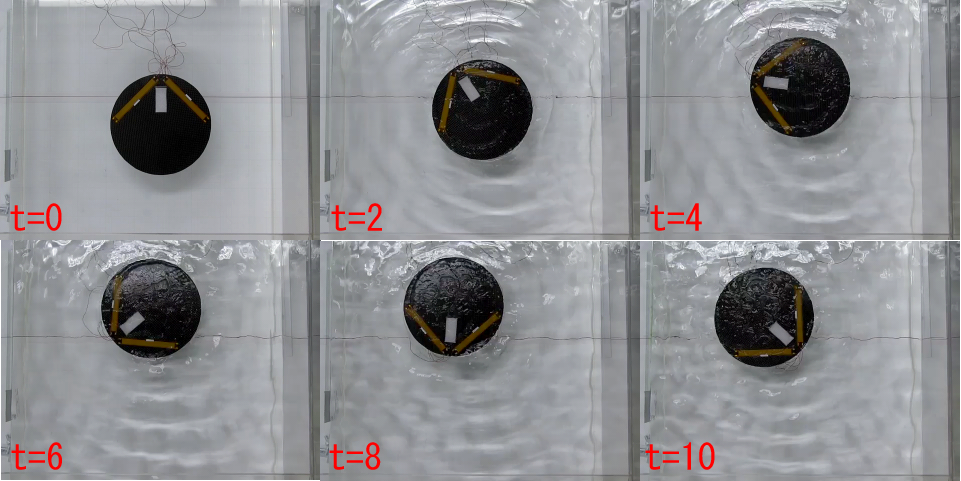}}
    \caption{Rotational movement}
    \label{hidari}
  \end{center}
\end{figure}

Rotational movement was observed at frequency of 5Hz and 6Hz with 75deg phase difference, as shown in Fig.\ref{hidari}.
The direction of the movement was reversed when the phase difference was reversed to $-$75deg.
When the phase difference was 0deg, forward movement appeared.
For other phase differences, forward turning movements were observed.

\section{Analysis and discussion}
\subsection{Parameter determination using beam model}
The experimental results indicated that most observed movements were placed on low frequency area, and they appeared at particular frequency.
It can be said that there is possibility of driving using eigenmodes.
Therefore, we tried to investigate the propulsion principle of those movements, based on structural vibration modes.
In order to do this, eigenmode and its frequency of robots in fluid are investigated.
We used ANSYS, an FEM software for simulation.

\begin{figure}[H]
  \begin{center}
    \resizebox{85mm}{!}{\includegraphics{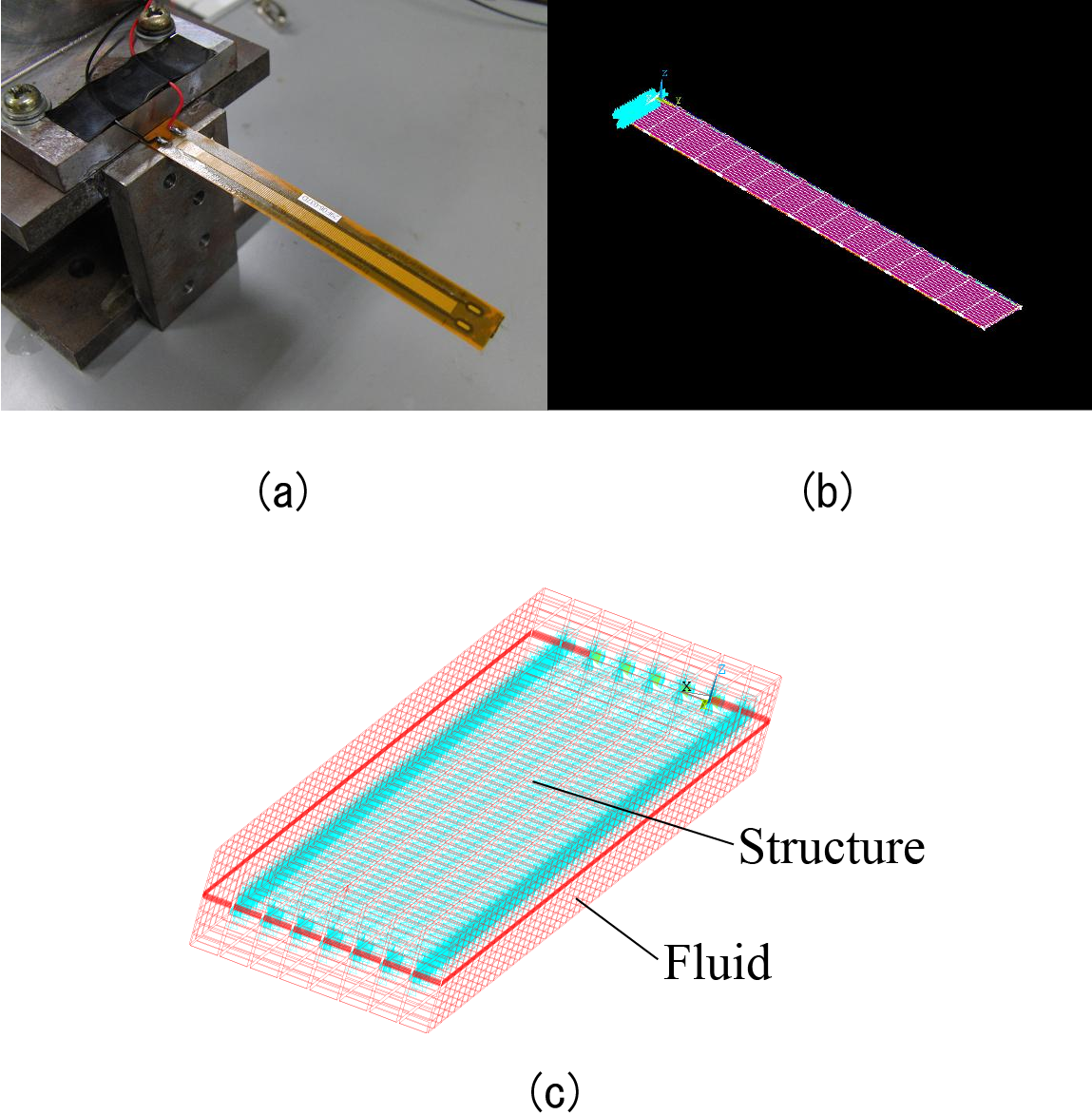}}
    \caption{Piezoelectric beam and its model}
    \label{beam}
  \end{center}
\end{figure}

First we made a basic beam model to determine parameters.
Fig.\ref{beam}(a) shows the beam (96.0$\times$12.4$\times$0.55mm) which consists of pizoelectic fiber composite, carbon plate and epoxy adhesive.
Fig.\ref{beam}(b) is the model of the beam by FEM.
To consider influence of fluid, we also made structural-fluid model, shown in Fig.\ref{beam}(c).
This model consists of structural elements and fluid elements, and both of elements are coupled as acoustic analysis.

\begin{figure}[H]
  \begin{center}
    \resizebox{85mm}{!}{\includegraphics{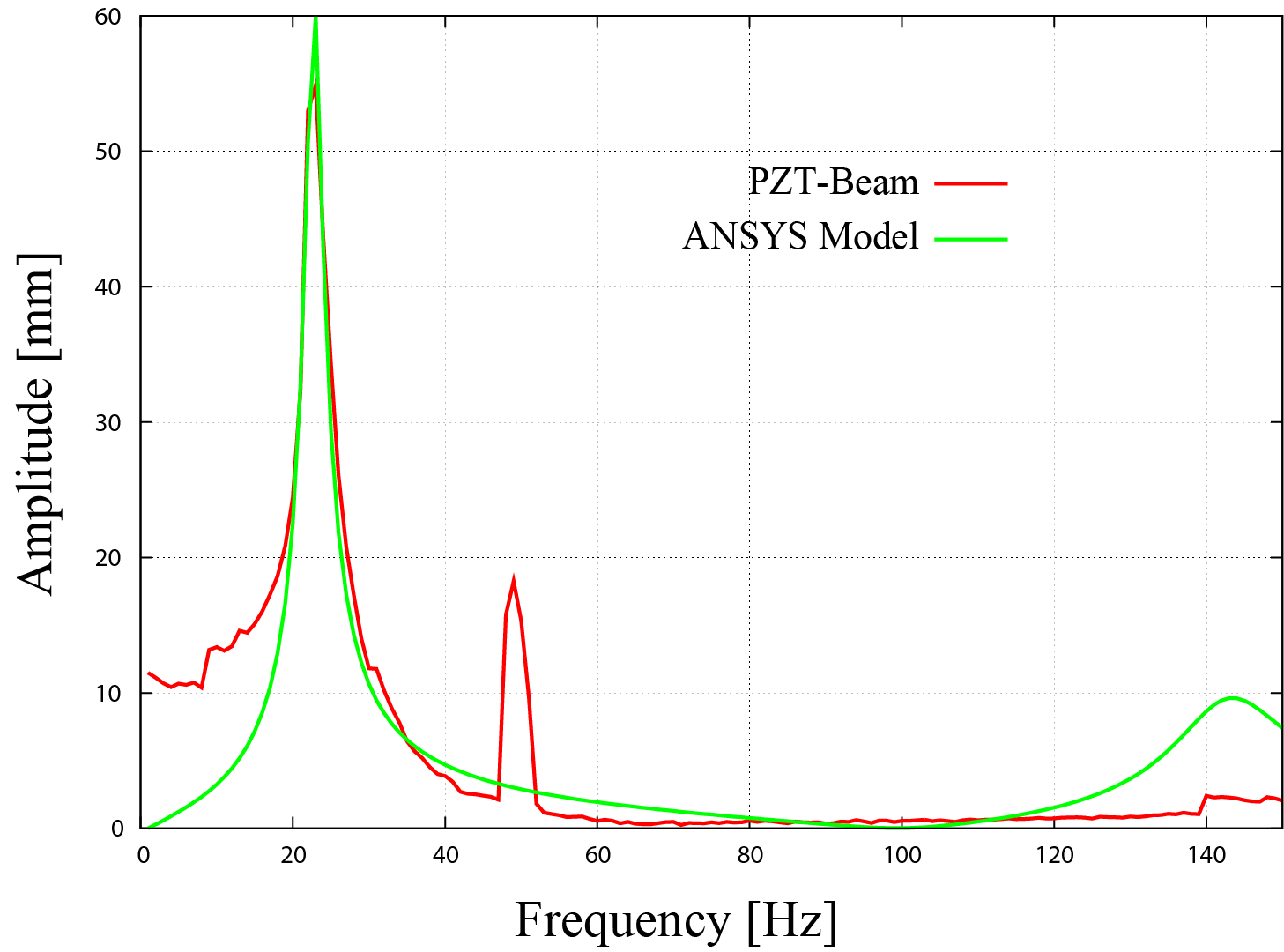}}
    \caption{Frequency response results in air}
    \label{air}
  \end{center}
\end{figure}

\begin{figure}[H]
  \begin{center}
    \resizebox{85mm}{!}{\includegraphics{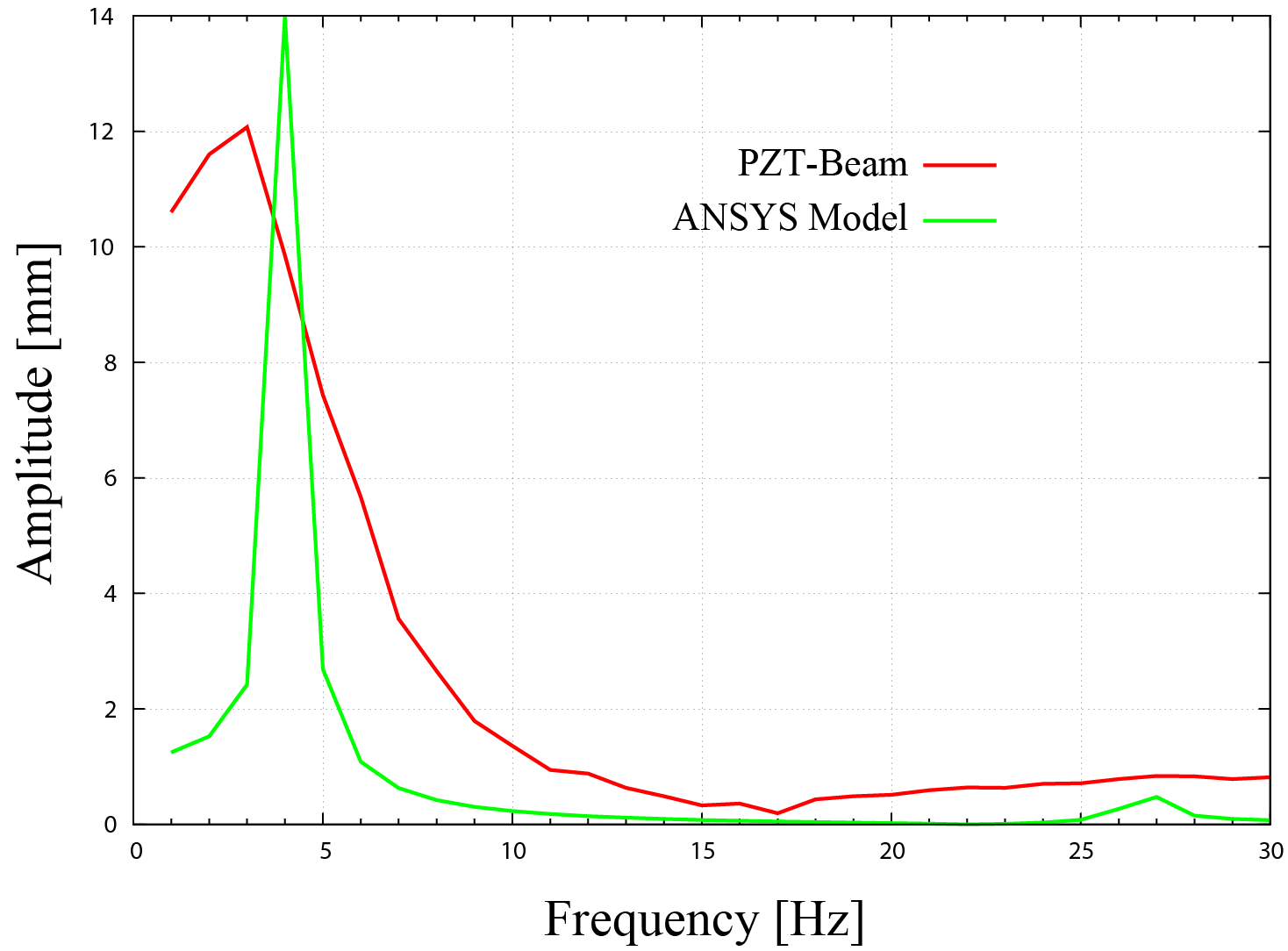}}
    \caption{Frequency response results in fluid}
    \label{fluid}
  \end{center}
\end{figure}

Fig.\ref{air} and Fig.\ref{fluid} are results of frequency response about the beam and the model.
We used a laser displacement gauge for measuring the beam displacement.
Tab.3 shows the comparison of measured eigenfrequency and calculated eigenfrequency.
Tab.4 shows the material properties of the beam and they are same as the robots, and thickness of these material are approximate measured value.
The property of the insulated fluid we used is, 1830kg/m$^3$ for the density, 600m/s for the acoustic velocity.

\begin{table}[H]
\begin{center}
\vspace{-1mm}
\caption{Comparison of eigenfrequencies}
\begin{tabular}{|c|c|c|c|c|}
\hline
~~ & \multicolumn{2}{|c|}{Frequency in air (Hz)} & \multicolumn{2}{|c|}{Frequency in fluid (Hz)}\\
\hline
Order & \makebox[4em]{1} & 2 & \makebox[4em]{1} & 2 \\
\hline
Beam & 23 & 141 & 3 & 27 \\
\hline
Model & 22.7 & 142.4 & 4.2 & 26.6 \\
\hline
\end{tabular}
\end{center}
\end{table}

\begin{table}[H]
\begin{center}
\caption{Material properties}
\vspace{1mm}
\begin{tabular}{|c|c|c|c|}
\hline
Properties & Carbon plate & PZT fiber & Adhesive \\
\hline
Density (kg/m$^3$) & 1643 & 4750 & 1140 \\
\hline
Elastic modulus (GPa)& 20.5 & 15.9 & 3.2 \\
\hline
Poisson's ratio & 0.3 & 0.31 & 0.34 \\
\hline
Thickness (mm) & 0.2 & 0.3 & 0.05 \\
\hline
\end{tabular}
\end{center}
\end{table}

\subsection{Robot models}
\begin{figure}[H]
  \begin{center}
    \resizebox{85mm}{!}{\includegraphics{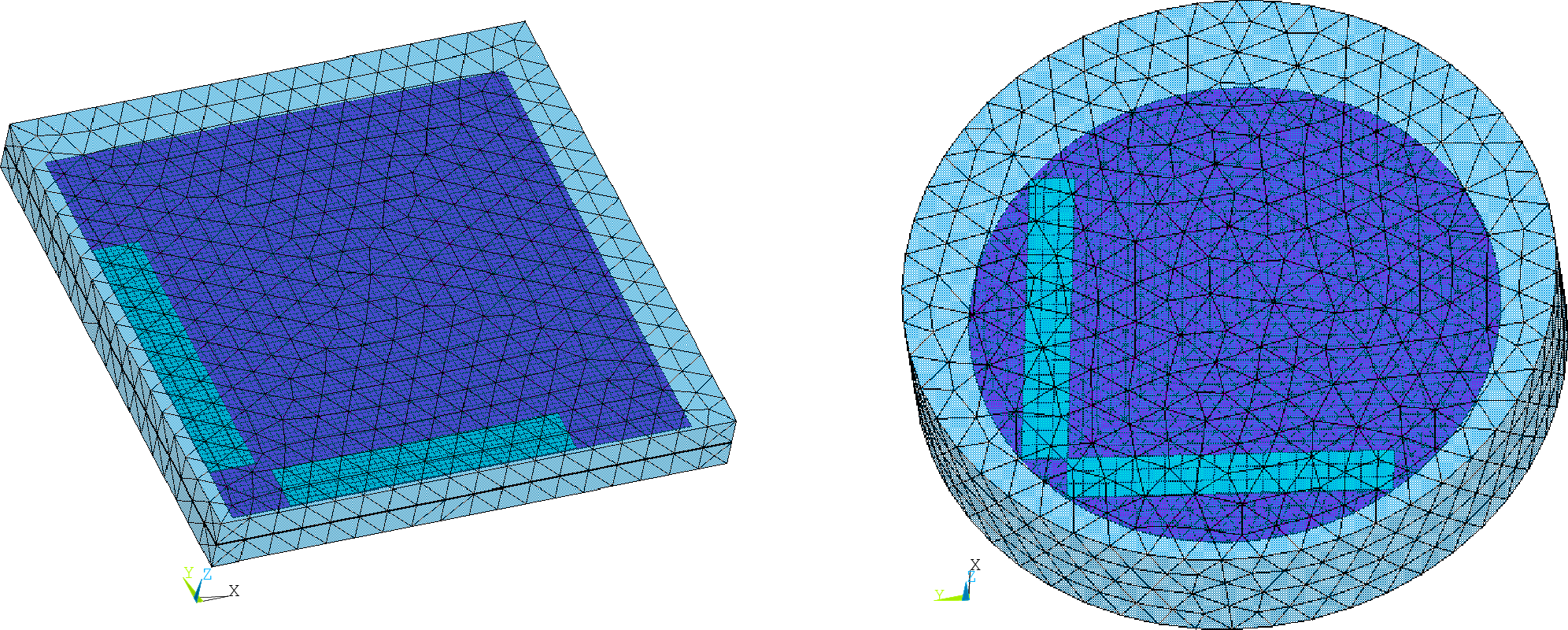}}
    \caption{Robot models}
    \label{robotmodels}
  \end{center}
\end{figure}

Fig.\ref{robotmodels} shows the models of the robots. The models consist of structural elements and fluid elements using determined parameters.

\subsection{Discussion on movements}
The calculated eigenfrequency by simulation of the robots are shown in Tab.5 and Tab.6.

\begin{table}[H]
\begin{center}
\caption{Eigenfrequency of rectangular robot}
\vspace{1mm}
\begin{tabular}{|c|c|c|c|c|c|}
\hline
Order & 1 & 2 & 3 & 4 & 5 \\
\hline
Frequency (Hz) & 1.38 & 2.35 & 2.50 & 4.80 & 5.55 \\
\hline
Order & 6 & 7 & 8 & 9 & 10 \\
\hline
Frequency (Hz) & 7.00 & 7.02 & 10.00 & 10.68 & 12.18 \\
\hline
\end{tabular}
\end{center}
\end{table}

\begin{table}[H]
\begin{center}
\caption{Eigenfrequency of circular robot}
\vspace{1mm}
\begin{tabular}{|c|c|c|c|c|c|}
\hline
Order & 1 & 2 & 3 & 4 & 5 \\
\hline
Frequency (Hz) & 1.87 & 2.06 & 3.02 & 5.23 & 5.31 \\
\hline
Order & 6 & 7 & 8 & 9 & 10 \\
\hline
Frequency (Hz) & 7.98 & 8.44 & 10.58 & 11.19 & 15.92 \\
\hline
\end{tabular}
\end{center}
\end{table}

In these tables, degenerate modes appear as combination of near two frequencies.
Then, we consider the movements of the rectangular shape robot as follows.

\subsubsection{Rotational movement}
The rotational movement was observed at 5Hz.
Therefore, from Tab.5, it is considered that order of corresponding modes are 4th and 5th.
Fig.\ref{matome}(a) shows the 4th mode. In this figure, different colors mean distribution of structural deformation: "red" is upward, "blue" is downward, and "green" is middle.
When the phase difference is 0deg, structural deformation is balanced as shown in Fig.\ref{matome}(b), and fluid-flows are also balanced or too little to generate a movement.
Therefore, the robot is static.
When the phase difference is 90deg, structural deformation is changed as a degenerate mode, shown as Fig.\ref{matome}(c).
In this case fluid-flows are changed their directions, and contour of flows shifted from center of mass. As a result, a moment for rotation was generated and rotational movement appeared.
In other cases, forward turning movements appeared since fluid-flows are unbalanced like Fig.\ref{matome}(d).

\subsubsection{Backward movement}
The backward movement was observed at 6Hz.
It is considered that order of corresponding modes are 6th and 7th. Fig.\ref{matome}(e) shows the 6th mode.
When the phase difference is 0deg, at the head structural deformation is enhanced by actuation, where fluid-flow is greater than back like Fig.\ref{matome}(f).
Therefore, backward movement appeared.
When the phase difference is 15deg, structural deformation is changed as a degenerate mode, and backward turning movement appeared as shown in Fig.\ref{matome}(g).
This case is similar to the rotational movement.
When the phase difference is 180deg, structural deformation as shown in Fig.\ref{matome}(h) appeared and forward movement appeared.
In this case, deformation of the head is small due to asynchronous actuation.

\subsubsection{Translational movements}
\begin{figure}[H]
  \begin{center}
    \resizebox{85mm}{!}{\includegraphics{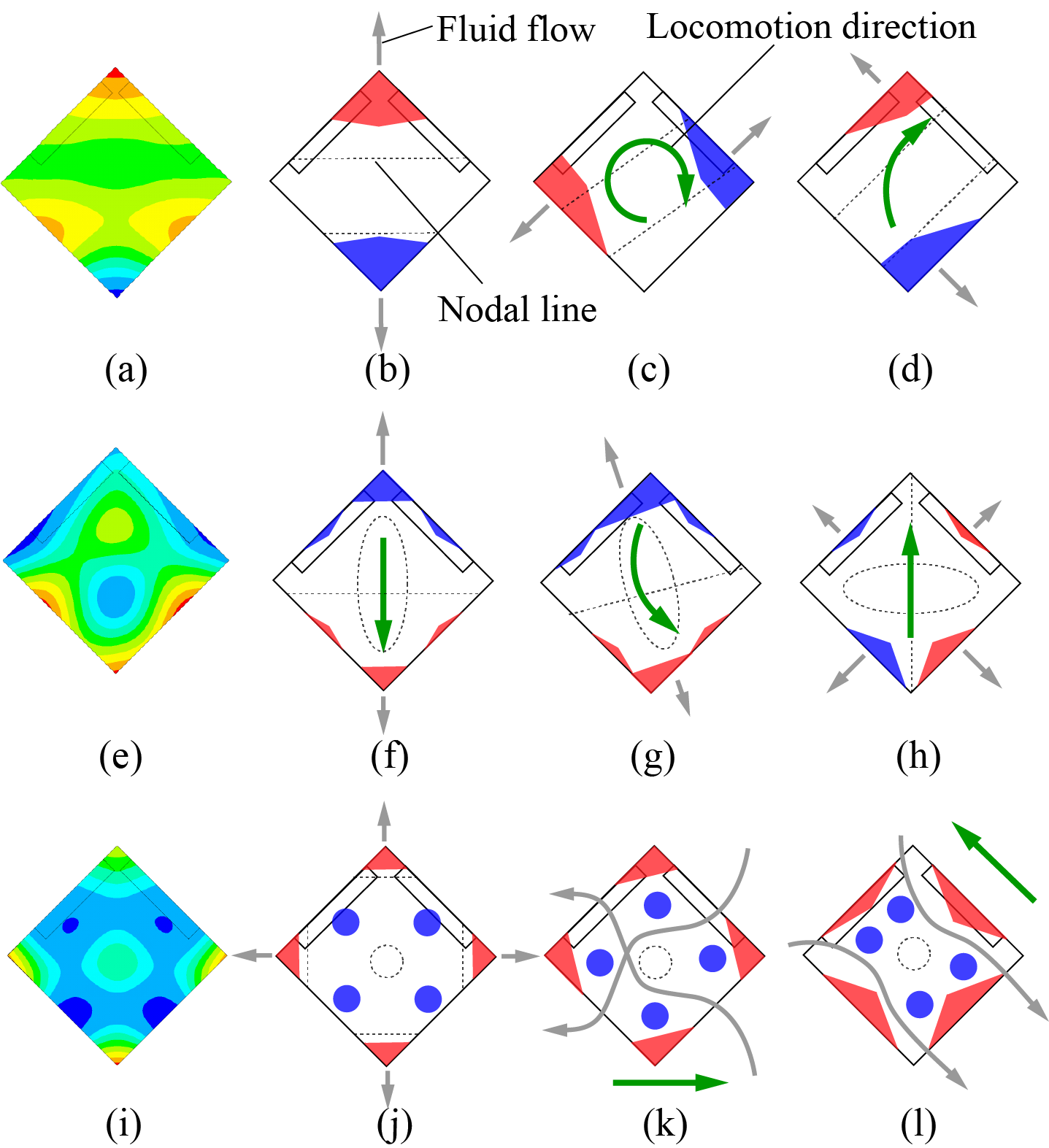}}
    \caption{Modes and estimated structural deformations}
    \label{matome}
  \end{center}
\end{figure}

The translational movement was observed at 8Hz.
It is considered that orders of corresponding modes are 8th and 9th. Fig.\ref{matome}(i) shows the 6th mode.
When the phase difference is 0deg, structural deformation is balanced as shown in Fig.\ref{matome}(j), and fluid-flows are also balanced or too little to generate a movement.
Therefore, the robot is static. This case is similar to the rotational movement.
Because of this mode has complex structural deformation, it is difficult to estimate the movements clearly.
It is considered that there is phenomenon same as rotational and backward movement at each phase differences such as 60deg and 120deg, shown as Fig.\ref{matome}(k) and Fig.\ref{matome}(l).
When the phase difference is 180deg, there is also phenomenon similar to the case of backward movement. As a result, forward movement appeared.

The reason why movements were not observed at other frequencies is that, a low frequency can not make sufficient speed of the structural deformation to generate an effective fluid-flow.
On the other hand, a high frequency has sufficient speed but has not sufficient structural displacement to generate an effective fluid-flow.
We thought that the backward movement by changing voltage range to $+1000$V$\sim+1500$V
is not the result of the particular eigenmode because this movement is distributed to certain frequency area,
but it is considered that there is particular structural deformation since this movement was also observed on the circular robot with same condition.
And it is also considered that to realize the movements, it needs some unbalance in robot shape.
A circle is symmetric to any direction but rectangular is not.
That is the reason why the circular robot has no movements on particular frequency.
From the results, it can be considered that the phase difference is the factor for controlling the degenerate modes.

\section{Conclusion}
In this paper, we proposed a novel propulsion method of flexible underwater robots.
The prototypes based on the method has been developed,
and unique movements such as movements with multiple DOFs, holonomic and non-holonomic movements have been demonstrated successfully.
The experimental results indicated that the possibility of driving using eigenmodes and was also shown by FEM analysis result.
Further work will be done on the quantitative analysis about the relation between movement and structural modes for designing such flexible robots.

\end{document}